\documentclass{article}

\usepackage{PRIMEarxiv}

\usepackage[utf8]{inputenc} 
\usepackage[T1]{fontenc}    
\usepackage{hyperref}       
\usepackage{url}            
\usepackage{booktabs}       
\usepackage{amsfonts}       
\usepackage{nicefrac}       
\usepackage{microtype}      
\usepackage{lipsum}
\usepackage{fancyhdr}       
\usepackage{graphicx}       
\graphicspath{{media/}}     

\title{Fine-Tuning or Fine-Failing? Debunking Performance Myths in Large Language Models}

\author{
  Scott Barnett, Zac Brannelly, Stefanus Kurniawan, Sheng Wong \\
  Applied Artificial Intelligence Institute\\
  Deakin University \\
  Geelong\\
  \texttt\{{scott.barnett, zac.brannelly, stefanus.kurniawan, sheng.wong\}@Deakin.edu.au} \\
}

\begin{document}
\maketitle

\begin{abstract}
Large Language Models (LLMs) have the unique capability to understand and generate human-like text from input queries. When fine-tuned, these models show enhanced performance on domain-specific queries. OpenAI highlights the process of fine-tuning, stating: "To fine-tune a model, you are required to provide at least 10 examples. We typically see clear improvements from fine-tuning on 50 to 100 training examples, but the right number varies greatly based on the exact use case." This study extends this concept to the integration of LLMs within Retrieval-Augmented Generation (RAG) pipelines, which aim to improve accuracy and relevance by leveraging external corpus data for information retrieval. However, RAG's promise of delivering optimal responses often falls short in complex query scenarios. This study aims to specifically examine the effects of fine-tuning LLMs on their ability to extract and integrate contextual data to enhance the performance of RAG systems across multiple domains. We evaluate the impact of fine-tuning on the LLMs' capacity for data extraction and contextual understanding by comparing the accuracy and completeness of fine-tuned models against baseline performances across datasets from multiple domains. Our findings indicate that fine-tuning resulted in a decline in performance compared to the baseline models, contrary to the improvements observed in standalone LLM applications as suggested by OpenAI. This study highlights the need for vigorous investigation and validation of fine-tuned models for domain-specific tasks. 
\end{abstract}


\section{Introduction}
In recent years, the advancements in Large Language Models (LLMs) have given the ability for machines to comprehend and generate human-like responses based on input queries. These LLMs, such as GPT, Mixtral, and Gemini, have demonstrated proficiency in a wide range of linguistic tasks, such as sentiment analysis, text generation, and conversational agents \cite{yang2023harnessing}. They possess the ability to apply learned knowledge to a wide range of areas, even those not specifically covered during their training. This capacity for generalisation allows them to adapt to new contexts and challenges beyond their initial training environments. Consequently, the capabilities of LLMs allow for their swift deployment in various tasks without the need for additional training data.

However, LLMs suffer from hallucination, a phenomenon where these models generate plausible-sounding but irrelevant or nonsensical information when asked domain-specific questions \cite{yao2023llm, yang2023harnessing}. This phenomenon is particularly concerning in fields where accuracy is paramount. Further, LLMs may sometimes provide inaccurate answers due to limitations in training data or inherent biases due to its internal representation. These limitations of LLMs highlight the need for adopting more nuanced approaches in improving their generative capabilities, leading to more contextually accurate and reliable outputs in various specialised domains.

Retrieval-augmented generation (RAG) serves as a promising solution by merging retrieval mechanisms with the generative capabilities of LLMs to synthesize contextually relevant, accurate, and up-to-date information \cite{lewis2021retrievalaugmented}. This integration allows RAG systems to enhance LLM responses by leveraging domain-specific knowledge bases or datasets, thereby improving the accuracy and relevance of answers without the need for further training. This framework significantly reduces development time by eliminating the need for extensive knowledge graph creation and detailed data curation and cleaning. However, RAG still faces multiple limitations, as highlighted by this study \cite{barnett2024seven}.

One potentially effective strategy for addressing the challenge encountered by LLMs with RAG pipeline in delivering accurate and relevant information to users is through fine-tuning pretrained LLM models. This process involves training existing pre-trained LLMs on domain-specific curated data, thereby enhancing their answering capabilities by adjusting the weights of the model's parameters. Fine-tuning involves optimizing the model's performance for the target task by adjusting and updating its weights based on domain-specific data during training. This process allows the model to learn task-specific information, thereby improving its ability to generate accurate and relevant responses. Fine-tuning is essential for adapting pre-trained LLMs to new tasks or domains without requiring complete retraining from scratch, which results in improved cost efficiency and reduced computational overhead.

In this study, we examine the effects of fine-tuning LLMs within a RAG pipeline on their question-answering performance. Initially conducted on private and proprietary datasets from a telecommunications company consisting of customer inquiries to the support team, we replicated the experiment using publicly available datasets. This research seeks to answer two research questions: Firstly, how do fine-tuned models compare to its baseline counterparts in a RAG pipeline? Secondly, does the size of the training dataset have an impact on the effectiveness of fine-tuning? By addressing these questions, we aim to contribute to a better understanding of how fine-tuning affects the question-answering abilities of RAG-integrated LLMs across various domains.

\section{Related Works}

One potentially effective strategy for addressing the challenge encountered by LLMs with RAG pipeline in delivering accurate and relevant information to users is through fine-tuning pretrained LLM models. This process involves training existing pre-trained LLMs on domain-specific curated data, thereby enhancing their answering capabilities by adjusting the weights of the model's parameters. Fine-tuning involves optimizing the model's performance for the target task by adjusting and updating its weights based on domain-specific data during training. This process allows the model to learn task-specific information, thereby improving its ability to generate accurate and relevant responses. Fine-tuning is essential for adapting pre-trained LLMs to new tasks or domains without requiring complete retraining from scratch, which results in improved cost efficiency and reduced computational overhead.

The process of fine-tuning generalised AI models has been a long-standing practice across various tasks and domains for better predictive performance \cite{salehi2023study, bharadiya2023convolutional}. Similarly, in the field of generative AI, there has been a trend where numerous studies have introduced specific LLMs tailored to aid in diverse functions in various domains. The projected benefits of fine-tuning include a)lower cost, b) better representation of domain technology, c) improved instruction following and d) improved accuracy. These fine-tuned LLMs have served in specific field such as finance \cite{DBLP:journals/corr/abs-1908-10063,zhang2024enhancing,yang2023fingpt, shah2023zero}, medicine \cite{singhal2022large,wu2023pmc,li2023chatgpt}, \cite{10.1093/bioinformatics/btz682}, creative writing \cite{wang2024weaver}, climate \cite{webersinke2022climatebert} and law \cite{cui2023chatlaw,nguyen2023brief}. 

These fine-tuned LLMs showed improved capabilities compared to the general LLMs in specific tasks. For instance, In the medical field, Med-PaLM originally proposed by Google, is able to surpass generalised LLMs in answering medical questions with better scientific consensus, comprehension, reasoning capabilities and completeness when evaluated \cite{singhal2022large}. In some evaluations such as information retrieval, it even comes close to human clinicians with 2\% differences. In the field of creative writing, a group of researchers developed Weaver, a series of LLMs based on open-source models, demonstrated that their fine-tuned models better across three categories: style, relevance, and creativity \cite{wang2024weaver} with other LLMs. 

In the legal domain, ChatLaw, enhanced with an additional self-attention mechanism and fine-tuned on Chinese legal datasets, demonstrated a significant reduction in hallucination compared to other generalised LLMs with a higher ELO score \cite{cui2023chatlaw}. In finance, the RoBERTa, fine-tuned on multiple financial dataset, performed better tone classification, sentiment analysis and named entity recognition tasks when compared to some of the generalised LLMs. Another study by JP Morgan Research Group observed similar results in financial text analysis \cite{li2023chatgpt}. Various fine-tuned LLMs such as outperformed generalised LLMs such as GPT4 entity extraction and sentiment analysis \cite{li2023chatgpt}. In recommendation tasks, the fine-tuned LLaMA7B model, leveraging the TALLRec framework and trained on 100 examples, exceeded the performance of general LLMs like GPT-4 in movie and book recommendations \cite{Bao_2023}. 

\section{Experimental Setups}

\subsection{Datasets}
The study utilised three open-source question-answering datasets to investigate how fine-tuning influences the performance of RAG-based LLM systems in delivering accurate and relevant responses. This approach ensures an extensive evaluation across different fields, providing a diverse understanding of how these systems perform in various domains. For each question in the datasets, the system fetches the top five relevant chunks from the vector store. These chunks are identified based on their semantic similarity to the input question, using a vector similarity search algorithm. The selected chunks are then concatenated to form a comprehensive RAG prompt. The RAG prompt serves as an augmented context for the question-answering system, enhancing GPT-4's ability to generate more accurate and contextually relevant answers. The datasets used in this study were:

\begin{enumerate}
    \item \textbf{BioASQ:} The biomedical dataset consists of 15000 documents and 1000 question and answer pairs constructed from a group of biomedical experts. 
    \item \textbf{Natural Questions (NQ):} The dataset is a large-scale dataset for open-ended question answering. It comprises queries of Google searches from real users and corresponding answers that are associated with Wikipedia page. 
    \item \textbf{Qasper:} is a dataset comprising 5,049 questions derived from 1,585 published Natural Language Processing (NLP) papers. Each question was formulated by an NLP practitioner who had access only to the title and abstract of the corresponding paper. Subsequently, other NLP practitioners provided answers to these questions along with supporting evidence.

\end{enumerate}

\subsection{Models \& Fine-tuning}
In this study, we used three models: Mistral \cite{jiang2024mixtral}, LlaMA2 \cite{touvron2023llama}, and GPT-4 \cite{openai2024gpt4}. Mistral and LlaMA2 are open-source, while GPT-4 is proprietary. We fine-tuned Mistral and LlaMA2 on sets of 200, 500, and 1000 question-answer pairs from each dataset to explore how training size affects performance. These sets of questions are not included in the final evaluation. As benchmarks, we used the base versions of all three models without fine-tuning. Performance for each dataset was then assessed across the non-fine-tuned models and the fine-tuned versions. All tested models, including the base and fine-tuned versions of Mistral and LlaMA2, were integrated within a RAG pipeline, which utilised the dataset (excluding data used for fine-tuning) reserved for fine-tuning, to ensure a fair assessment across all datasets. The models Mistral and LlaMA2 were fine-tuned using varying configurations to optimize performance. The hardware used for fine-tuning consisted of Intel Xeon Platinum 8452Y processors, with configurations supporting up to four NVIDIA H100 GPUs or eight A100 GPUs, and 500GB of memory. Key configurations tested included:
\begin{itemize}
    \item \textbf{Number of Epochs:} Variations in the number of training epochs to find the optimal duration for model convergence.
    \item \textbf{Effective Batch Size:} Adjustments in batch size and gradient accumulation steps to balance training efficiency and computational resource usage.
    \item \textbf{Efficiency Techniques:} Evaluation of LoRa versus QLoRa to determine which method offers superior model efficiency without compromising output quality.
    \item \textbf{LoRa Hyperparameters:} Tuning of rank and alpha parameters within the LoRa settings to refine the adaptation process.
\end{itemize}

\subsection{Evaluation}
To evaluate the performance of both base and fine-tuned models, we employed a custom version of the G-Evals framework, which utilizes Large Language Models (LLMs) with a chain-of-thought (CoT) approach and a form-filling paradigm \cite{liu2023geval}. In this experiment, Mistral were utilised as the LLM for assessing the baseline and fine-tuned models. This framework assesses text output quality by comparing it against human judgments based on defined metrics, providing scores that reflect a human-like understanding of the answer quality.

We used two main evaluation metrics: accuracy and completeness. The accuracy metric assesses how well the generated answer addresses the question, its relevance and the sufficiency of supporting evidence, and the absence of irrelevant or misleading information. The completeness metric evaluates the extent to which an answer covers the topics and details posed by the question.

The model judge were given instruction to rate responses using a 5-point scoring system for both evaluation metrics. For accuracy, a score of 1 indicates "Poor Accuracy," and a score of 5 indicates "Excellent Accuracy." Similarly, for completeness, a score of 1 denotes "Incomplete or Irrelevant" output, while a score of 5 signifies a "Complete and Comprehensive" response. To mitigate variance in the scoring, the evaluation process using the model judge was repeated 10 times. The final score was determined by averaging these results, ensuring a more consistent and reliable evaluation result.

\section{Result} 

\begin{figure}
    \centering
    \includegraphics[width=1\linewidth]{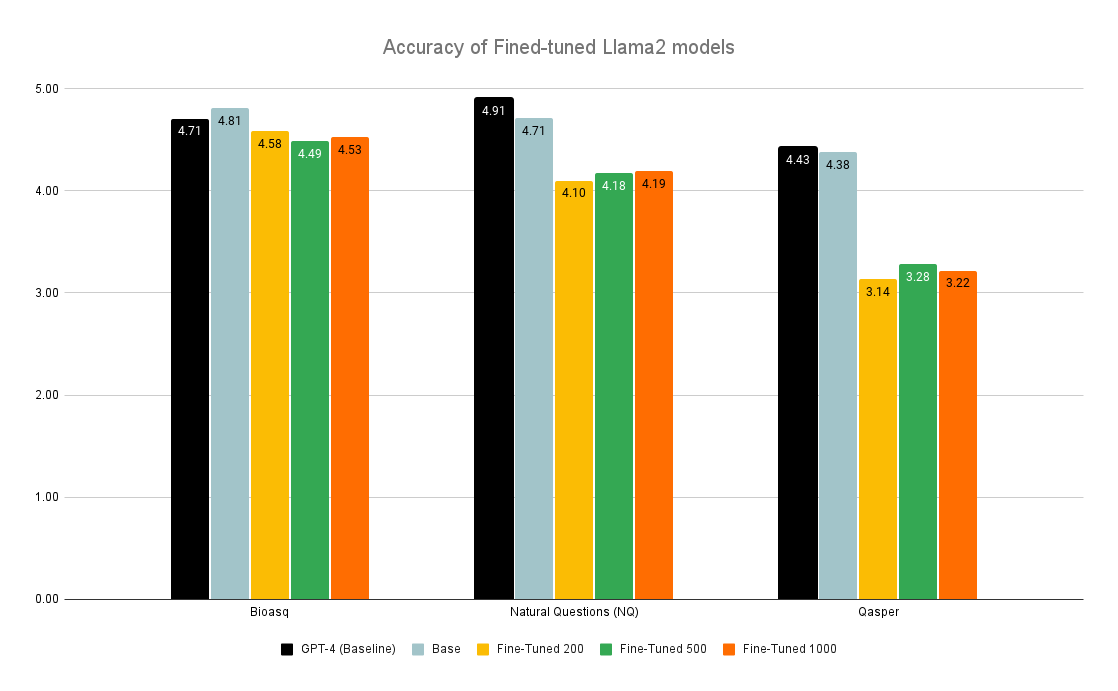}
    \caption{Comparisons of accuracy for fine-tuned Llama2 models and baseline models across three datasets.}
    \label{fig:acc Llama2}
\end{figure}

\begin{figure}
    \centering
    \includegraphics[width=1\linewidth]{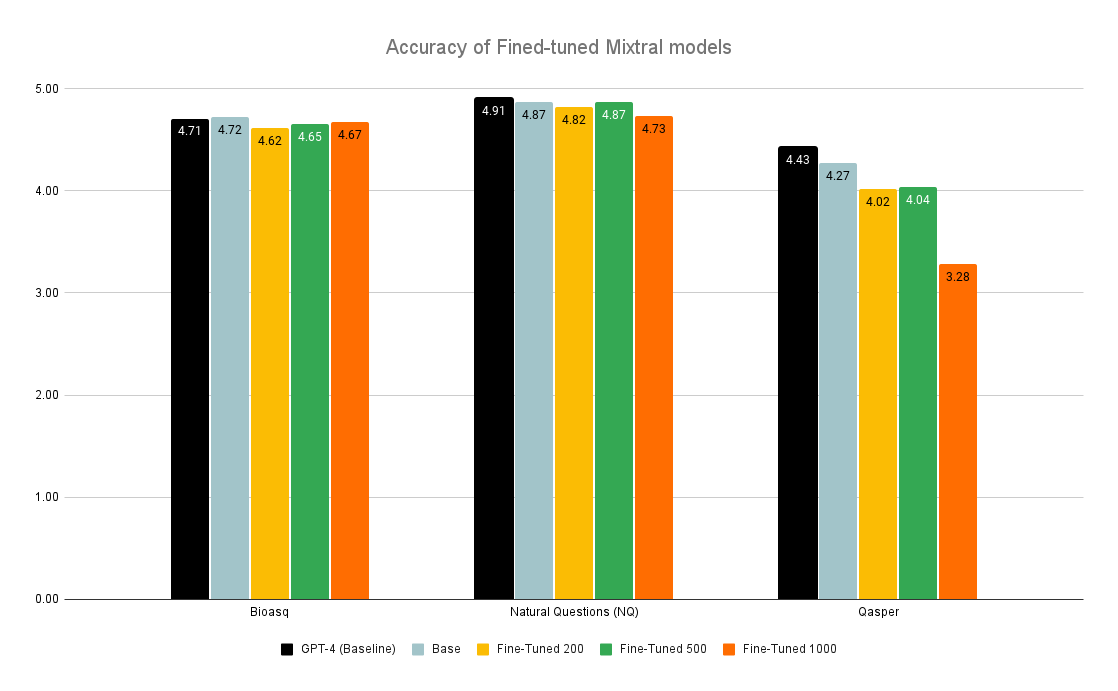}
    \caption{Comparisons of accuracy for fine-tuned Mixtral models and baseline models across three datasets.}
    \label{fig:acc Mixtral}
\end{figure}

\begin{figure}
    \centering
    \includegraphics[width=1\linewidth]{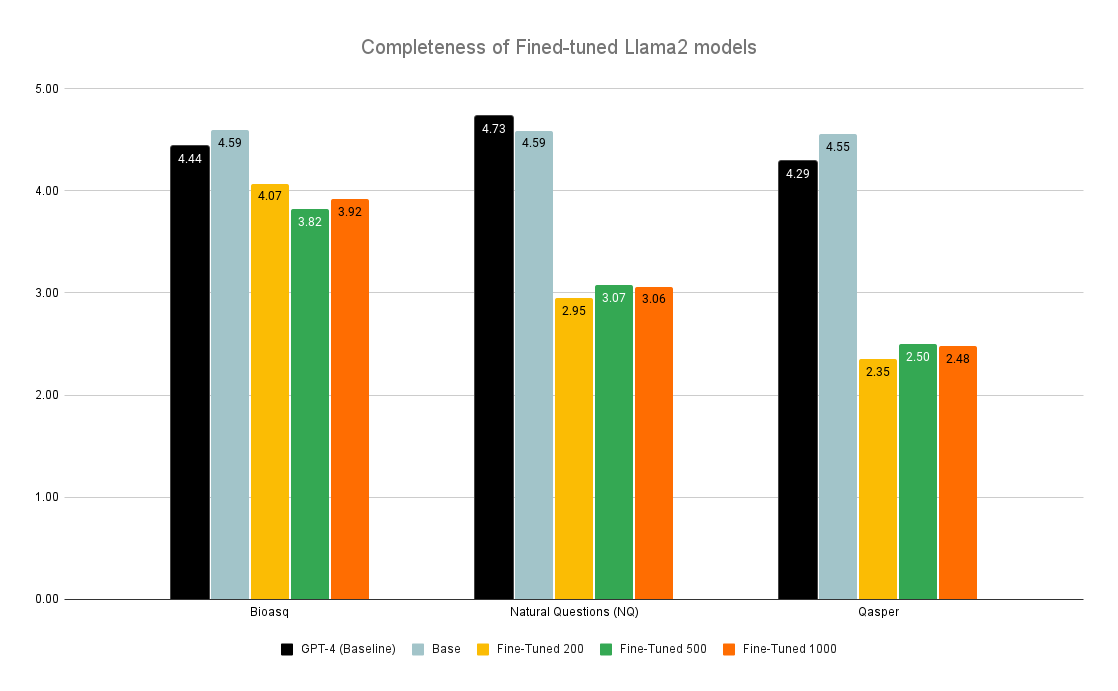}
    \caption{Comparisons of completeness for fine-tuned Llama2 models and baseline models across three datasets.}
    \label{fig:completeness Llama2}
\end{figure}

\begin{figure}
    \centering
    \includegraphics[width=1\linewidth]{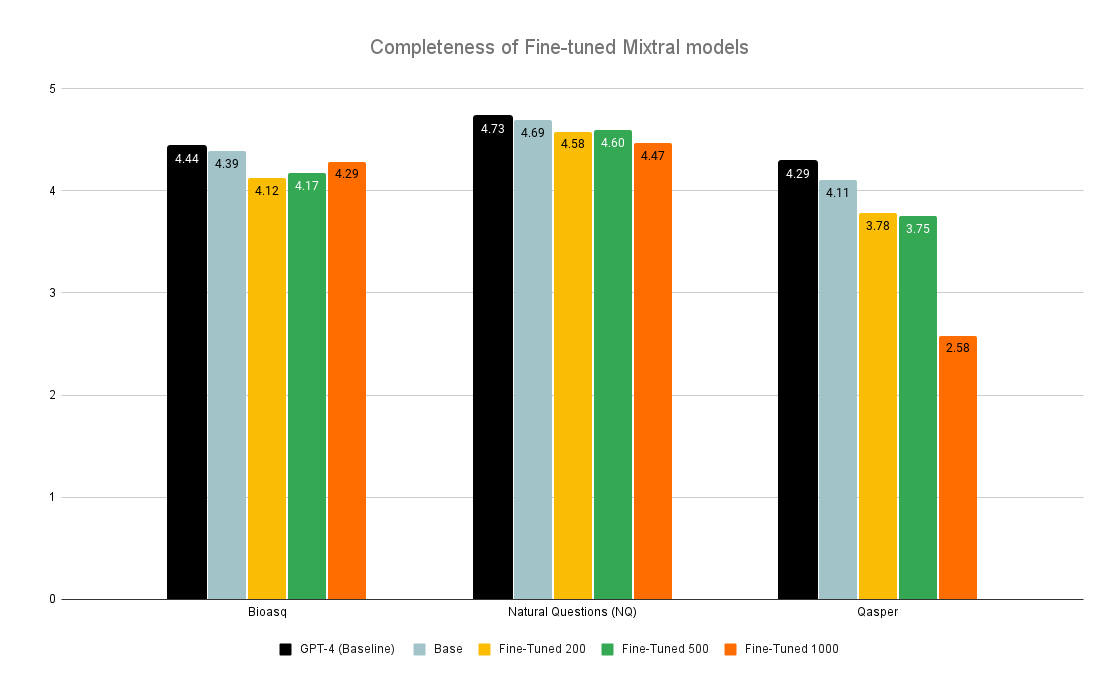}
    \caption{Comparisons of completeness for fine-tuned Mixtral models and baseline models across three datasets.}
    \label{fig:completeness Mixtral}
\end{figure}

In this study, we evaluated our fine-tuning models using two evaluation metrics: accuracy and completeness. Figure \ref{fig:acc Llama2} and \ref{fig:acc Mixtral} present the accuracy of the fine-tuned models compared to the baseline models across three datasets. The base models of Mixtral and Llama2 without fine-tuning outperformed any of their fine-tuned counterparts across all datasets, except for the NQ dataset. The Mixtral model fine-tuned with 500 samples performed identically to its baseline counterpart, achieving an accuracy score of 4.87. Similar observations were made with the baseline GPT-4 model, which performed better than any of the fine-tuned models across all datasets.

Figure \ref{fig:completeness Llama2} and Figure \ref{fig:completeness Mixtral} illustrate the completeness of the fine-tuned models compared to the baseline models across different datasets. The base models of Llama2 and Mixtral without fine-tuning outperformed any of their fine-tuned counterparts across all datasets. Similarly, the baseline GPT-4 model achieved a higher completeness score than any of the fine-tuned models.

\section{Discussion}
Based on the results presented in the previous section, it is evident that the baseline Mixtral and Llama2 models surpassed their fine-tuned iterations across all datasets evaluated. Nevertheless, it is noteworthy that the performance gap was marginal for certain models and datasets; specifically, the accuracy of the BioASQ dataset for both fine-tuned Mixtral and Llama2 models and the accuracy of the Mixtral model on the NQ dataset remained close to their respective baselines. Similarly, fine-tuned Mixtral models performed marginally worse than the baseline model. 

A significant decline in performance was observed with the Qasper dataset for both models, where the accuracy score decreased by more than one point on the scale. All fine-tuned Llama2 models and the Mixtral model fine-tuned at the 500 level saw a decrease in accuracy exceeding one score point. In terms of completeness, several fine-tuned models exhibited a reduction of two score points, indicating a significant impact on performance post-fine-tuning. Additionally, it was observed that fine-tuning the Llama2 models from their baseline configuration resulted in a more significant decline in performance compared to the fine-tuning of Mixtral models from their respective baselines. 
Llama2 experienced a sharp decline in accuracy, dropping from 4.38 to 3.14, and completeness, which fell from 4.55 to 2.35, on the fine-tuned dataset of 200 samples — the lowest score achieved across any models. This indicates that fine-tuning tends to worsen performance issues more severely in Llama2 models than in Mixtral models.

In this study, we also evaluated fine-tuned models using different sample sizes. Contrary to expectations, fine-tuned models often performed worse than the baseline models. Moreover, in some cases, fine-tuning with increasingly larger sample size deteriorated both the accuracy and completeness of the model’s answering capabilities. Within the Qasper dataset, expanding the sample size used for fine-tuning from 200 to 500, and then to 1000 samples, resulted in decreased performance metrics. Specifically, when the sample size increased from 500 to 1000, the accuracy for the Mixtral model experienced a sharp decline from 4.04 to 3.28, and completeness dropped from 3.75 to 2.58. In other instances, there was no significant relationship between the sample size and the performance of the fine-tuned models.

Hence, our experiments highlight that fine-tuning does not equate to better accuracy or completeness, and fine-tuning on large domain-specific datasets harms the ability of LLMs to provide accurate and complete answers when integrated within RAG pipeline. Numerous studies have advocated for fine-tuning as a method to enhance performance in domain-specific tasks, suggesting it as a superior approach compared to using generalized large language models (LLMs). However, our findings align with some existing research, such as the studies by \cite{kalyan2023survey, kuzman2023chatgpt, li2023chatgpt}, which also observed poorer performance on fine-tuned models despite not incorporating a RAG pipeline.

\section{Limitations}
The study faced several limitations that could be addressed in future research. Firstly, the dataset size used for training was fairly small, and a larger dataset might yield different results. Secondly, the study explored only a limited number of hyperparameter configurations for our models, which was constrained by the availability of hardware resources. Further, due to hardware resource limitations, training was performed on single nodes, which led to smaller batch sizes and could have affected the training results. Thirdly, the evaluation method involved the use of an LLM via G-Evals to assess performance, a technique that could be considered less reliable than other methods involving human evaluation. Additionally, different prompts and retrieval mechanisms were employed within the RAG system. This introduced variables that could have influenced the performance of the fine-tuned models. Lastly, the context size provided to the LLMs was limited to five chunks of 1,500 characters, which may have inflated the performance of smaller models due to their potentially better handling of smaller context sizes.

\section{Conclusion}
In this study, we evaluated the effectiveness of fine-tuning (less than 1000 samples) on various publicly open datasets for answer generation. The experiments were replicated from our internal experiment conducted using a private telecommunication dataset. We observed that fine-tuning LLMs within a RAG pipeline negatively impacts their performance in answer generation. Our findings hopefully provide another lens on the nuanced impact of fine-tuning on LLMs and suggest that while fine-tuning can enhance model performance in specific scenarios, it may not be universally advantageous. This study contributes by presenting instances where fine-tuning does not lead to the expected improvements, contrasting with the majority of the studies where fine-tuning is typically shown to be beneficial. Future studies could explore fine-tuning increased sample sizes (more than 1000) to validate our findings.

\section{Acknowledgments}
We would like to thank Telstra Corporation, Steve Morris and Sara Keretna for providing the support and datasets required to complete this unit of work. 

\bibliographystyle{unsrt}  
\bibliography{references}

\end{document}